\documentclass{article}

\usepackage{microtype}
\usepackage{graphicx}
\usepackage{subfigure}
\usepackage{booktabs} % for professional tables
\usepackage{subcaption}
\usepackage{tabularx} % Add this in your preamble

\usepackage{hyperref}

\usepackage[accepted]{icml2025}
\usepackage{amsmath}
\usepackage{amssymb}
\usepackage{mathtools}
\usepackage{amsthm}

\theoremstyle{plain}

\theoremstyle{definition}

\theoremstyle{remark}

\icmltitlerunning{Mixup Augmentation for Audio}

\begin{document}

\twocolumn[
\icmltitle{Bridging the Language Gap: Synthetic Voice Diversity via Latent Mixup for Equitable Speech Recognition}

\begin{icmlauthorlist}
\icmlauthor{Wesley Bian}{ucla}
\icmlauthor{Xiaofeng Lin}{ucla}
\icmlauthor{Guang Cheng}{ucla}
\icmlaffiliation{ucla}{University of California Los Angeles, Department of Statistics, Los Angeles, United States of America}
\end{icmlauthorlist}
\icmlcorrespondingauthor{Xiaofeng Lin}{bernardo1998@g.ucla.edu}
\icmlcorrespondingauthor{Wesley Bian}{wbian@g.ucla.edu}
\icmlkeywords{Machine Learning, ICML, ASR, Audio, Synthetic Data, Data Augmentation, Mixup, Fairness}

\vskip 0.3in

]


\printAffiliationsAndNotice{\icmlEqualContribution} % otherwise use the standard text.

\begin{abstract}

Modern machine learning models for audio tasks often exhibit superior performance on English and other well-resourced languages, primarily due to the abundance of available training data. This disparity leads to an unfair performance gap for low-resource languages, where data collection is both challenging and costly. In this work, we introduce a novel data augmentation technique for speech corpora designed to mitigate this gap. Through comprehensive experiments, we demonstrate that our method significantly improves the performance of automatic speech recognition systems on low-resource languages. Furthermore, we show that our approach outperforms existing augmentation strategies, offering a practical solution for enhancing speech technology in underrepresented linguistic communities.

\end{abstract}
\section{Introduction and Related Work}

Automatic Speech Recognition (ASR) and other voice-related machine learning technologies have made remarkable advances in recent years, largely propelled by the abundance of data available for English. However, this data imbalance has led to a significant diversity gap: ASR systems perform well for English but often struggle with the world’s other 7,000+ languages, perpetuating inequities in access to advanced speech technologies. Addressing this disparity is essential to ensure that the benefits of modern machine learning are distributed equitably across all linguistic communities.

Data augmentation for speech recognition and synthesis has been explored at multiple levels of granularity, including \emph{waveform}, \emph{acoustic feature}, and \emph{latent representation}. Classic perturbative methods---additive noise, speed or tempo modification, vocal--tract--length warping, and channel convolution---improve robustness, but do not explicitly enhance the diversity of speaker characteristics represented in the dataset. SpecAugment~\citep{park2019specaugment}, which introduced time-warping and frequency/time masking on log-Mel spectrograms, remains a strong baseline.

Recent work has shown that interpolation in latent spaces can yield smoother decision boundaries and improved generalization. Mixup~\citep{zhang2018mixup} and Manifold Mixup~\citep{verma2019manifold} demonstrate the benefits of linear interpolation between hidden states in vision models, while MixRep~\citep{xie2023mixrep} adapts this approach to ASR by mixing encoder activations, achieving gains on low-resource English corpora. Latent Filling~\citep{bae2023latentfilling} applies interpolation to speaker embeddings in zero-shot TTS, enhancing similarity without requiring additional data collection.

Style-based augmentation, particularly through voice conversion models that disentangle speaker characteristics from linguistic content, has further expanded the potential for generating synthetic diversity. Models such as CycleGAN-VC and StarGAN-VC enable non-parallel, many-to-many voice transformations, though audible artifacts in generated audio can limit their downstream utility~\citep{vca}. Despite their use in augmentation pipelines, there remains room for improvement.

To the best of our knowledge, no prior work has explored the application of mixup \emph{within} the latent code space of style encoders. We propose \textsc{LatentVoiceMix}, a method that operates on this intermediate representation, preserving phonetic structure while expanding the latent convex hull associated with each language. Empirically, we demonstrate that mixup in the style-encoder space yields superior performance compared to existing augmentation methods, controlling for the amount of synthetic audio generated. By bridging latent interpolation theory with codec-level modeling, our approach introduces the first fairness-oriented synthetic data generator at the style layer and provides new evidence that latent convexity is critical for multilingual speech learning.

\section{Methodology}

We adapt the Diff-HierVC voice conversion model~\citep{diffhiervc} to generate synthetic speech data with novel speaker characteristics while preserving the original linguistic content. The core of our approach is to disentangle and manipulate the speaker timbre and linguistic information in audio samples, leveraging a diffusion-based architecture for high-fidelity voice conversion.

\subsection{Voice Conversion Model Overview}

The Diff-HierVC model separates an input audio file into two distinct representations: (1) the linguistic content, corresponding to the words spoken, and (2) the speaker timbre, which encapsulates the unique, non-linguistic characteristics of a person's voice. Speaker timbre refers to the qualities that make a voice unique, independent of linguistic factors such as accent or language. The model enables the recombination of the linguistic encoding from a source audio file with the speaker timbre encoding from a different, target audio file, thereby synthesizing speech that retains the content of the source but adopts the vocal characteristics of the target speaker.

\subsection{Data Augmentation Procedure}

Our augmentation pipeline proceeds as follows:

\begin{enumerate}
    \item \textbf{Audio Cleaning:} All audio files in the input dataset are first denoised using the \texttt{noisereduce} Python package~\citep{noisereduce} to ensure high-quality input for subsequent processing.
    
    \item \textbf{Speaker Timbre Extraction and Storage:} 
    For each audio file in the corpus, we apply the encoding module of Diff-HierVC to extract a fixed-length speaker timbre representation. Specifically, the style encoder produces a 255-dimensional vector that characterizes the unique, time-invariant vocal attributes of each speaker, independent of linguistic content~\citep{diffhiervc}. These timbre vectors are systematically stored on the filesystem, enabling efficient retrieval and reuse throughout the augmentation pipeline.
    
    \item \textbf{Source Selection:} For each data point, the audio file is designated as the \emph{source}, providing the linguistic content for augmentation.
    
    \item \textbf{Target Selection:} A separate audio file, spoken by a different speaker, is randomly selected from the dataset to serve as the \emph{target}.
    
    \item \textbf{Mixup Timbre Selection:} An additional, pre-saved speaker timbre, distinct from both the source and target speakers, is randomly selected to facilitate mixup.
    
    \item \textbf{Voice Conversion with Mixup:} The source linguistic encoding, obtained from the Diff-HierVC model, is combined with a convex combination of the target and mixup speaker timbres. Specifically, let $\mathbf{t}_{\mathrm{target}}$ and $\mathbf{t}_{\mathrm{mixup}}$ denote the timbre vectors of the target and mixup speakers, respectively. The mixed timbre vector is computed as
    \[
        \mathbf{t}_{\mathrm{mixed}} = \lambda\, \mathbf{t}_{\mathrm{target}} + (1-\lambda)\, \mathbf{t}_{\mathrm{mixup}},
    \]
    where $\lambda \sim \mathrm{Beta}(\alpha, \beta)$ with $\alpha = 0.5$ and $\beta = 0.5$. The model then generates a synthetic audio file that retains the original linguistic content but exhibits a novel, realistic-sounding speaker timbre.
    
    \item \textbf{Post-processing:} The synthesized audio is further cleaned using \texttt{noisereduce} to remove any residual artifacts.
    
    \item \textbf{Transcript Assignment:} The transcript associated with the synthetic audio is inherited directly from the original source file, as the linguistic content remains unchanged.
\end{enumerate}

\subsection{Benefits and Impact}

This augmentation strategy enables the creation of an expanded and more diverse voice corpus without the need for additional data collection. By generating realistic synthetic voices with preserved linguistic content, our method significantly enhances the diversity of training data available for Automatic Speech Recognition (ASR) systems. Empirical results demonstrate that incorporating this augmented data leads to meaningful improvements in ASR model performance, particularly for underrepresented speaker profiles.

\section{Experiments and Results}

We conducted a series of experiments to evaluate the effectiveness of our proposed mixup augmentation technique for improving Automatic Speech Recognition (ASR) performance, and to compare its impact with other established augmentation methods. All experiments report Word Error Rate (WER), a standard metric for ASR evaluation in which lower values indicate better performance.

\subsection{Datasets}

We evaluated the effectiveness of our mixup augmentation technique using three speech corpora: a Wolof dataset~\citep{wolof}, the CSTR VCTK corpus for English~\citep{vctk}, and the an4 dataset for additional low-resource experimentation~\citep{an4}. The Wolof corpus consists of 16 hours of transcribed speech from 14 speakers, representing a low-resource language spoken in West Africa. The VCTK corpus provides approximately 44 hours of English speech from 109 speakers with diverse accents, while the an4 dataset is a small English corpus commonly used for benchmarking ASR pipelines in resource-constrained settings.

\subsection{ASR Models}

To rigorously assess the impact of our augmentation method, we trained and evaluated two widely adopted automatic speech recognition (ASR) frameworks: Whisper~\citep{whisper} and NVIDIA NeMo~\citep{nemo}. These models were selected due to their prevalence in both academic and industrial ASR research, as well as their support for multilingual and low-resource scenarios. This experimental design enables us to demonstrate the practical benefits and generalizability of our approach across diverse languages and model architectures.

\subsection{Alternative Baseline Augmentation Methods}

To assess the effectiveness of our proposed mixup augmentation strategy, we conducted comparative experiments against several widely used audio data augmentation techniques. \emph{Waveform augmentation} directly manipulates raw audio signals using time-stretching, amplitude scaling, and pitch shifting, applied randomly to each file with the \texttt{audiomentations} library~\citep{audiomentations}. \emph{Spectrogram augmentation}, as in SpecAugment~\citep{park2019specaugment}, increases data diversity by masking random frequency bands or time intervals in the spectrogram representation. \emph{Voice conversion augmentation} generates new samples by transferring the linguistic content of an audio file to the vocal characteristics of a different speaker~\citep{speakerAugmentation}, thereby increasing speaker diversity while preserving the original transcript.

\subsection{AN4: Low-Resource English}

We first trained the NVIDIA NeMo ASR model from scratch for 50 epochs on the AN4 dataset. We compared four training regimes: no augmentation, waveform augmentation, voice conversion augmentation, and our mixup augmentation. For all augmentation methods, the dataset size was increased by 33\%. Table \ref{tab:an4Experiment} shows that mixup augmentation significantly improves the performance of this model, and is superior to traditional waveform augmentation. 

\begin{table}[h]
\centering
\caption{ASR Performance (WER) on AN4 with Different Augmentation Methods}
\label{tab:an4Experiment}
\begin{tabular}{lcc}
\toprule
\textbf{Training Data} & \textbf{Augmentation} & \textbf{WER} \\
\midrule
AN4 only & None & 0.785 \\
AN4 + 33\% & Waveform & 0.436 \\
AN4 + 33\% & Voice conversion & 0.424 \\
AN4 + 33\% & Mixup & \textbf{0.339} \\
\bottomrule
\end{tabular}
\end{table}

\subsection{Wolof vs. English: Addressing Language Bias}

To simulate a realistic multilingual training scenario, we constructed a dataset by randomly sampling 8 hours of Wolof speech from the original corpus, ensuring equal representation from each speaker. For English, we similarly sampled 24 hours of data from the VCTK corpus. The NeMo ASR model was then trained for 50 epochs on this combined dataset. As shown in Table~\ref{tab:multilingual}, the model achieved substantially better performance on English than on Wolof, highlighting a persistent bias toward the higher-resource language.

We then augmented the Wolof portion of the dataset with an additional 16 hours of synthetic data generated using our mixup augmentation technique, and repeated the training. The results demonstrate a substantial reduction in the performance gap between Wolof and English, indicating improved fairness and inclusivity.

\begin{table}[h]
\centering
\caption{WER for Multilingual ASR on Wolof and English}
\label{tab:multilingual}
\begin{tabularx}{\columnwidth}{Xccc}
\toprule
\textbf{Training Data} & \textbf{Wolof} & \textbf{English} & \textbf{Gap} \\
\midrule
8h Wolof + 24h English & 0.796 & 0.562 & 0.234 \\
24h Wolof (aug.) + 24h English & 0.725 & 0.550 & \textbf{0.175} \\
\bottomrule
\end{tabularx}
\end{table}

\subsection{Finetuning Whisper on Wolof: Comparison with Other Augmentation Methods}

In practice, many automatic speech recognition (ASR) systems are developed by fine-tuning large pretrained models, such as Whisper by OpenAI, rather than training from scratch. To assess the impact of our augmentation methods in this setting, we fine-tuned the \texttt{whisper-tiny} model on 8 hours of original Wolof data for 4 epochs. For each original sample, two synthetic samples were generated using various augmentation techniques, resulting in a tripled dataset size. We also evaluated the perceptual quality of the augmented data using the average SpeechMOS score~\citep{speechmos}, or, in the case of no augmentation, the original data. SpeechMOS is a neural model that predicts human-perceived speech quality on a scale from 1 to 5. For spectrogram augmentation, SpeechMOS was not computed, as this method does not generate waveform outputs.

Table~\ref{tab:whisper} summarizes the WER achieved with different augmentation strategies. Our mixup augmentation method consistently outperformed spectrogram augmentation, waveform augmentation, and conventional voice conversion augmentation, yielding the lowest WER.

\begin{table}[h]
\centering
\caption{WER for Whisper Finetuned on Wolof with Different Augmentation Methods}
\label{tab:whisper}
\begin{tabular}{lcc}
\toprule
\textbf{Augmentation Method} & \textbf{WER} & \textbf{SpeechMOS} \\
\midrule
None & 0.283 & 2.661\\
Spectrogram Augmentation & 0.242 & n/a\\
Waveform Augmentation & 0.217 & 2.117 \\
Voice Conversion Augmentation & 0.215 & 2.710\\
Mixup Augmentation (proposed) & \textbf{0.202} & 2.243\\
\bottomrule
\end{tabular}
\end{table}
\section{Analysis of Method}

\subsection{Ablation Study on Mixup Augmentation}

To understand the contribution of individual components within our mixup augmentation algorithm, we conducted an ablation study using 8 hours of original Wolof data for fine-tuning Whisper. The effectiveness of each variant was assessed using word error rate (WER) on the ASR task. Table~\ref{tab:ablation} summarizes the results for various configurations.

\begin{table}[h]
\centering
\caption{Ablation study of mixup augmentation variants on Wolof ASR (WER; lower is better).}
\label{tab:ablation}
\begin{tabularx}{\columnwidth}{l c}
\toprule
\textbf{Setting} & \textbf{WER} \\
\midrule
No Post-denoising, Source=Target (8h) & 0.235 \\
No Post-denoising, Source=Target (16h) & 0.221 \\
Mixup w/ 3 Speaker Timbres (16h) & 0.221 \\
No Post-denoising (16h) & 0.214 \\
Proposed Mixup (16h) & \textbf{0.202} \\
\bottomrule
\end{tabularx}
\end{table}

The results indicate that the full mixup algorithm, including post-denoising, achieves the lowest WER. Omitting the post-denoising step or altering the mixup configuration—such as using three timbres or setting the source and target to the same audio file—consistently results in higher error rates. These findings underscore the importance of post-denoising and careful design of the mixup process for optimal augmentation performance.

\subsection{Analysis of Speaker Timbre Distributions}

To further investigate the qualitative differences between augmentation methods, we performed principal component analysis (PCA) on the speaker timbre vectors of the 14 speakers in the Wolof corpus. We compared these to timbre vectors extracted from synthetic samples generated by both mixup and traditional waveform augmentation. As illustrated in Figure~\ref{fig:augmentation_comparison}, the speaker timbres produced by mixup are distributed more closely to those of the original speakers, whereas timbres from waveform augmentation exhibit greater variance and tend to lie outside the distribution of real speaker timbres. These results suggest that mixup augmentation generates synthetic data that not only increases training diversity but also more faithfully preserves the underlying structure of the original speaker timbre distribution. This property may explain the superior performance of mixup augmentation compared to traditional waveform augmentation.
\label{fig:augmentation_comparison}
\includegraphics[scale=0.5]{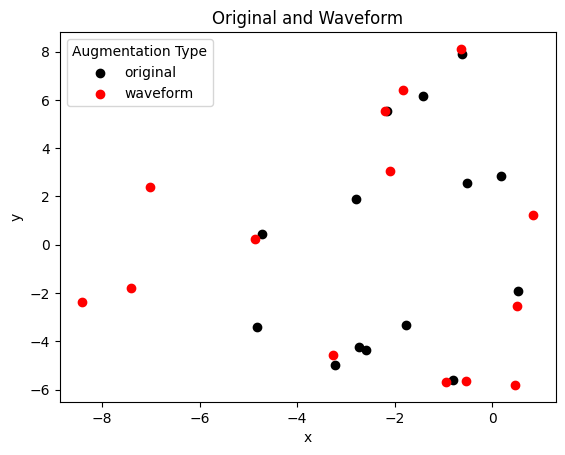}
\includegraphics[scale=0.5]{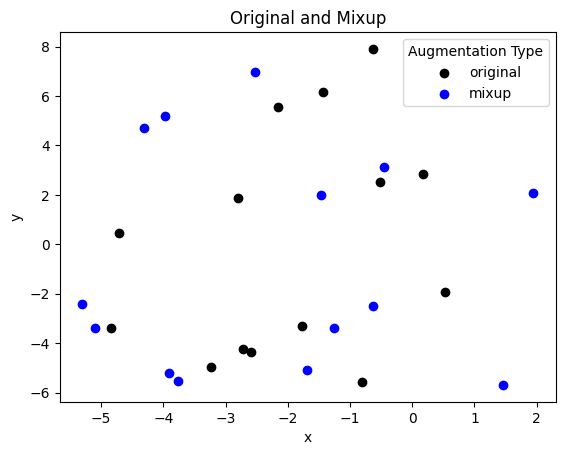}
\section{Conclusion}

We introduced a novel data augmentation technique that applies mixup in the latent space of voice conversion models, and demonstrated its effectiveness in enhancing Automatic Speech Recognition (ASR) performance. Our experiments show that this method outperforms traditional augmentation techniques, particularly for low-resource languages. These findings indicate that significant gains in ASR for underrepresented languages can be achieved without extensive data collection, promoting broader access to advanced speech technologies.

\section*{Impact Statement}

This work advances machine learning for audio in low-resource languages by reducing bias toward well-resourced languages. Our approach promotes equitable access to speech technology, enabling speakers of all languages to benefit from progress in machine learning systems.

\bibliography{custom}
\bibliographystyle{icml2025}

\end{document}